\documentclass[letterpaper, 10 pt, conference]{ieeeconf}  

\IEEEoverridecommandlockouts                              

\usepackage{siunitx}
\usepackage{graphicx}
\usepackage{mathrsfs}
\usepackage{array}
\usepackage{url}
\usepackage{bm}
\usepackage{comment}
\usepackage{mathrsfs}
\usepackage{amsfonts}
\usepackage{psfig}
\usepackage{graphics} 
\usepackage{epsfig} 
\usepackage{mathptmx} 
\usepackage{times} 
\usepackage{amsmath} 
\usepackage{amssymb}  
\usepackage{multicol} 
\usepackage{upgreek}
\usepackage{bm} 
\usepackage{cite}
\usepackage{subeqnarray}
\usepackage{cases}
\usepackage{multirow}
\usepackage{graphics} 
\usepackage{epsfig} 
\usepackage{mathptmx} 
\usepackage{times} 
\usepackage{booktabs}
\usepackage{float}
\usepackage{sansmath}
\usepackage{subcaption}
\usepackage{color}
\usepackage{relsize}
\usepackage{textcomp}

\newtheorem{definition}{\textbf{Definition}}

\newtheorem{theorem}{\rm\textbf{Theorem}}

\newtheorem{assump}{\rm\textbf{Assumption}}
\newtheorem{remark}{\rm\textbf{Remark}}
\usepackage{algorithm}
\usepackage{algpseudocode}

\title{\LARGE \bf

Safe Motion Planning and Control Using Predictive and Adaptive Barrier Methods for Autonomous Surface Vessels
}
\author{
Alejandro Gonzalez-Garcia, Wei Xiao, Wei Wang$^{\ast}$, Alejandro Astudillo, Wilm Decr\'e, \\ Jan Swevers, Carlo Ratti and Daniela Rus
\thanks{This work was supported by a grant from the Amsterdam Institute for Advanced Metropolitan Solutions (AMS) in Netherlands, and by the MIT-Belgium - KU Leuven Seed Fund from MIT International Science and Technology Initiatives (MISTI). A. Gonzalez-Garcia was supported by the Flanders Make SBO project ARENA (Agile \& Reliable Navigation). }
\thanks{A. Gonzalez-Garcia, A. Astudillo, W. Decr\'e and J. Swevers are with MECO Research Team, Department of Mechanical Engineering, KU Leuven, Belgium and with Flanders Make@KU Leuven, Belgium. {\tt\small \{alex.gonzalezgarcia, wilm.decre, jan.swevers\}@kuleuven.be}}
\thanks{W. Xiao and D. Rus are with the Computer Science and Artificial Intelligence Lab (CSAIL),  Massachusetts Institute of Technology, Cambridge, MA 02139 USA. {\tt\small \{weixy, rus\}@mit.edu}}%
\thanks{W. Wang is with the Marine Robotics Lab, Department of Mechanical Engineering, College of Engineering, University of Wisconsin-Madison, Madison, WI 53706 USA. {\tt\small \{wwang745\}@wisc.edu}}
\thanks{C. Ratti is with the SENSEable City Laboratory, Massachusetts Institute of Technology, Cambridge, MA 02139 USA. {\tt\small \{ratti\}@mit.edu}}
\thanks{$^{\ast}$Authors to whom correspondence may be addressed.}
}

\begin{document}

\maketitle
\thispagestyle{empty}
\pagestyle{empty}

\begin{abstract}
Safe motion planning is essential for autonomous vessel operations, especially in challenging spaces such as narrow inland waterways. However, conventional motion planning approaches are often computationally intensive or overly conservative. This paper proposes a safe motion planning strategy combining Model Predictive Control (MPC) and Control Barrier Functions (CBFs). We introduce a time-varying inflated ellipse obstacle representation, where the inflation radius is adjusted depending on the relative position and attitude between the vessel and the obstacle. The proposed adaptive inflation reduces the conservativeness of the controller compared to traditional fixed-ellipsoid obstacle formulations. The MPC solution provides an approximate motion plan, and high-order CBFs ensure the vessel's safety using the varying inflation radius. Simulation and real-world experiments demonstrate that the proposed strategy enables the fully-actuated autonomous robot vessel to navigate through narrow spaces in real time and resolve potential deadlocks, all while ensuring safety.

\end{abstract}

\section{Introduction}

Autonomous Surface Vessels (ASVs) have gained attention over the past few decades, and the demand is expected to increase as the marine industry continues to expand \cite{vagale2021}. Inland waterways, in particular, offer a sustainable mode of transporting goods and people, where ASVs can help increase safety and reduce operational errors \cite{Domenighini2024}. Beyond transportation, ASVs are also helpful in other applications such as hydrographic surveying, water quality monitoring, and waste removal \cite{Wei2023}. However, ASVs navigating inland waterways face unique challenges compared to coastal or marine scenarios \cite{Cheng2021}. The confined nature of narrow canals and rivers, combined with static and dynamic obstacles—such as other vessels—demands precise maneuvering capabilities \cite{shan2020, Streichenberg2023}. Therefore, developing safe motion planning and control systems is critical for successfully deploying ASVs in these settings. In this work, we consider a collision-free motion as a safe motion.

Model Predictive Control (MPC) has become a popular strategy for motion planning and control within the robotics community \cite{9802523,8678822,10777539}. 
The capabilities of handling multi-objective optimization, nonlinear dynamics, and system constraints have made MPC a powerful tool for integrated motion planning and control. However, challenges appear when multiple obstacles and complex shapes are considered. First, as the number of constraints increases, so does the computational complexity. This potentially endangers the real-time capabilities of the algorithm, as the solutions may not be able to meet predefined time requirements. MPC performs better with convex costs and constraint formulations, which lead to conventional ellipsoid representations of robots and obstacles. Although computationally cheap, common fixed ellipsoidal representations are overly conservative \cite{GONZALEZGARCIA2022oa}. In this work, we consider a conservative representation as one that sacrifices significant free space to achieve an efficient constraint formulation. This can lead to deadlocks, as the robots cannot find a feasible solution. Control Barrier Functions (CBFs) are optimization-based methods that have gathered attention for their safety guarantees and reduced computational needs \cite{Aaron2014,xiao2021high}. CBFs are based on Barrier Functions (BFs), which are Lyapunov-type functions, extended to formulate constraints for control systems. However, classic CBFs can only handle relative-degree-one systems, \textit{i.e.}, systems where the control input appears on the first derivative of the output. High Order CBFs (HOCBFs) \cite{xiao2021high} were proposed to create constraints for arbitrarily high relative degree systems. CBFs have high computational efficiency, since they can be implemented as a Quadratic Program (QP) when applied independently or with Control Lyapunov Functions \cite{Aaron2014}.




\subsection{Related Work}

Prior work for controlling ASVs that require precise maneuvering includes grid-based motion planners, which offer solutions that can account for obstacles and spatial boundaries, but generally do not consider vehicle dynamics or the feasibility of following the calculated path. In \cite{shan2020}, a Receding Horizon Planner (RHP) based on lexicographic search was applied to ASVs in urban waterways. Here, the RHP achieved better results than OpenPlanner \cite{Darweesh_2017jrm} and the Time Elastic Band (TEB) planner on cluttered environments. Although the RHP could operate online, its obstacle avoidance was overly conservative due to the fixed-radius inflation of the occupied pixels. Other approaches rely on local reactive planning, such as Velocity Obstacles \cite{Huang2019}, Artificial Potential Fields \cite{Shi2019}, or the Constant Avoidance Angle method \cite{Wiig2020}. Although these types of algorithms are usually computationally efficient, they are also conservative and may lead to deadlock situations in narrow or cluttered scenarios. Model Predictive Path Integral (MPPI) control was proposed in \cite{Streichenberg2023} to address multiple vehicles in urban waterways. MPPI leverages model knowledge, sampling techniques, and parallelization to compute a collision-free solution. While MPPI can handle non-convex or non-conservative problem formulations, its performance remains dependent on sampling quality and available computational resources. 

In contrast, MPC can efficiently address convex problems, although representing obstacles as circles or ellipses is conservative \cite{GONZALEZGARCIA2022oa}. Other approaches attempt to decrease the conservativeness by representing the vessel as a set of circles \cite{DeVries2022RegulationsCanals}. However, depending on the size of the boat, this strategy can still be overly conservative, and the computational complexity significantly increases as more constraints are required per obstacle. Non-conservative solutions can be achieved through polytope representations, although they often pose challenges in their formulation and handling, and more computational resources are needed compared to ellipsoidal representations. In \cite{Thirugnanam2022}, a method to formulate polytope-based constraints with discrete CBFs was proposed. The CBFs were defined as the collision constraints within an MPC framework, achieving successful performance even with complex shapes representing the robot and the environment. Still, constraint relaxations were applied to balance feasibility and safety. In \cite{Wei2024}, an MPC-CBF framework was proposed to control ASVs under disturbances robustly. Robust HOCBFs were designed to steer the boat within a boundary of the desired trajectory, considering an upper bound on the disturbances. The MPC provided an initial control reference, and the HOCBFs modified the control signal to add robustness. However, in contrast to the work in this manuscript, motion planning and obstacle avoidance were not addressed. At the core of this paper, we propose a new adaptive method to construct HOCBFs that maintains the efficiency of ellipsoid solutions but allows for a less conservative representation.

\subsection{Contributions}

This paper proposes an algorithm for safe motion planning for fully-actuated ASVs in narrow spaces with static and quasi-static floating obstacles. This work considers narrow spaces as tight spaces for the vessel to maneuver, which can only be tackled under specific orientations. A methodology is proposed to dynamically define obstacle inflation, reducing ellipsoid-based obstacle representations' conservativeness while maintaining computational efficiency. The algorithm leverages the MPC horizon to search for an approximate collision-free path, whereas High Order CBFs (HOCBFs) guarantee safety in tight spaces. The main contributions of this work include:\\
$\tiny {\bullet}$ An MPC-CBF framework for safe ASV collision avoidance through narrow spaces;\\
$\tiny {\bullet}$ HOCBFs with adaptive safety constraint design for  obstacle avoidance in narrow passages, and to remain within spatially safe boundaries;\\
$\tiny {\bullet}$ A safe deadlock recovery mechanism that leverages the full actuation configuration;\\ 
$\tiny {\bullet}$ Simulation and real-world experiments with a holonomic ASV, verifying the effectiveness of the MPC-CBF framework against multiple obstacles in narrow spaces.

\section{Preliminaries}

In this section, the ASV prototype and its dynamics are described, the problem formulation is addressed, and preliminaries on CBF/HOCBF theory are introduced.

\subsection{Quarterscale Roboat}


The Quarterscale Roboat \cite{WeiICRA2018} is an overactuated ASV with dimensions of 0.90 m in length $l$, 0.45 m in width $w$, and 0.15 m in height, and a weight of 15 kg. Four BlueRobotics T200 thrusters actuate the vessel, capable of holonomic 2D motion. The Roboat is equipped with a Velodyne VLP-16 LiDAR and a Microstrain 3DM-GX5-IMU. The onboard computer is an Intel NUC with Linux and Robot Operating System (ROS), and a STM32F103 auxiliary microprocessor. For more details, see \cite{WeiICRA2018}.

\subsection{ASV Dynamics}\label{ss:dynamic_model}

The ASV equations of motion are described by:
\begin{eqnarray}\label{MPCPredynamics}
&&\dot{\bm{\eta}}=\mathbf{R}(\bm{\eta})\mathbf{v} , \label{MFCPredynamicsA}\\
&&\bm{\tau}=\mathbf{M}\dot{\mathbf{v}}+\mathbf{C}(\mathbf{v})\mathbf{v}+\mathbf{D}(\mathbf{v})\mathbf{v} ,\label{MFCPredynamicsB}
\end{eqnarray}
where $\bm{\eta}=[x \quad y \quad \psi]^{T} \in\mathbb{R}^{3}$ is the position and orientation in the inertial reference frame, and $\mathbf{v}=[u \quad v \quad r]^{T}\in \mathbb{R}^{3}$ represents the vessel velocity in the body-fixed frame.  $\mathbf{R}(\bm{\eta})\in \mathbb{R}^{3\times3}$ is a transformation matrix, $\mathbf{M} \in \mathbb{R}^{3\times3}$ represents the added mass and inertia matrix, $\mathbf{C}(\mathbf{v})\in\mathbb{R}^{3\times3}$ stands for the Coriolis matrix, and $\mathbf{D}(\mathbf{v})\in\mathbb{R}^{3\times3}$ is the drag matrix. $\bm{\tau}=[\tau_u \quad  \tau_v \quad  \tau_r]^{T} \in\mathbb{R}^{3}$ contains the force and torque applied by the thrusters, defined by: 
\begin{eqnarray}\label{AppliedForceMaxtrix}
\bm{\tau}
=\mathbf{B}\mathbf{u}
=
\left[
 \begin{array}{cccc}
1                      &  1                                      &    0                       & 0\\
0                      &  0                                      &    1                       & 1\\
\dfrac{a_{\text{d}}}{2}&-\dfrac{a_{\text{d}}}{2}                 &     \dfrac{b_{\text{d}}}{2} &-\dfrac{b_{\text{d}}}{2}
\end{array}
\right]
 \left(
 \begin{array}{c}
f_1\\
f_2\\
f_3\\
f_4
\end{array}
\right),
\end{eqnarray}
where $\mathbf{B}\in\mathbb{R}^{3\times4}$ represents the control matrix, defining the arrangement of the thrusters. $\mathbf{u}=[f_1 \quad f_2 \quad f_3 \quad f_4]^{T}\in\mathbb{R}^{4}$ is the control vector, where $f_1$, $f_2$, $f_3$, and $f_4$ correspond to the forces generated by each thruster. $a_{\text{d}}$ is the distance between the port and starboard thrusters, and $b_{\text{d}}$ is the distance between the anterior and rear thrusters. See \cite{WeiICRA2018} for more details.

\subsection{Problem Formulation}

Consider a scenario where a predefined global path is given in terms of a parameterized curve $\pmb{p_d}(\omega) = [x_d(\omega), y_d(\omega)]^T$, where $\omega \in\mathbb{R}^+$ is the variable parameter. Then, the path-following goal is to remain close to the path, reducing the cross-track error 
    $y_e = -\sin(\gamma_p)(x-x_d(\omega^*)) + \cos(\gamma_p)(y-y_d(\omega^*))$,
and to keep the course of the ASV aligned with the path angle
    $\gamma_p = \mathrm{atan2}(y^\prime_d(\omega^*),x^\prime_d(\omega^*))$.
In these expressions, $\pmb{p_d}(\omega^*)$ is the path reference point. $\pmb{p_d}(\omega^*)$ is found by computing $\omega^*$, which minimizes the Euclidean distance between the ASV and the predefined path with $\min_\omega f(\omega)$, where
    $f(\omega) = (x-x_d(\omega))^2 + (y-y_d(\omega))^2.$

\begin{figure}[tb]
    \centering
    \includegraphics[width=0.6\linewidth] {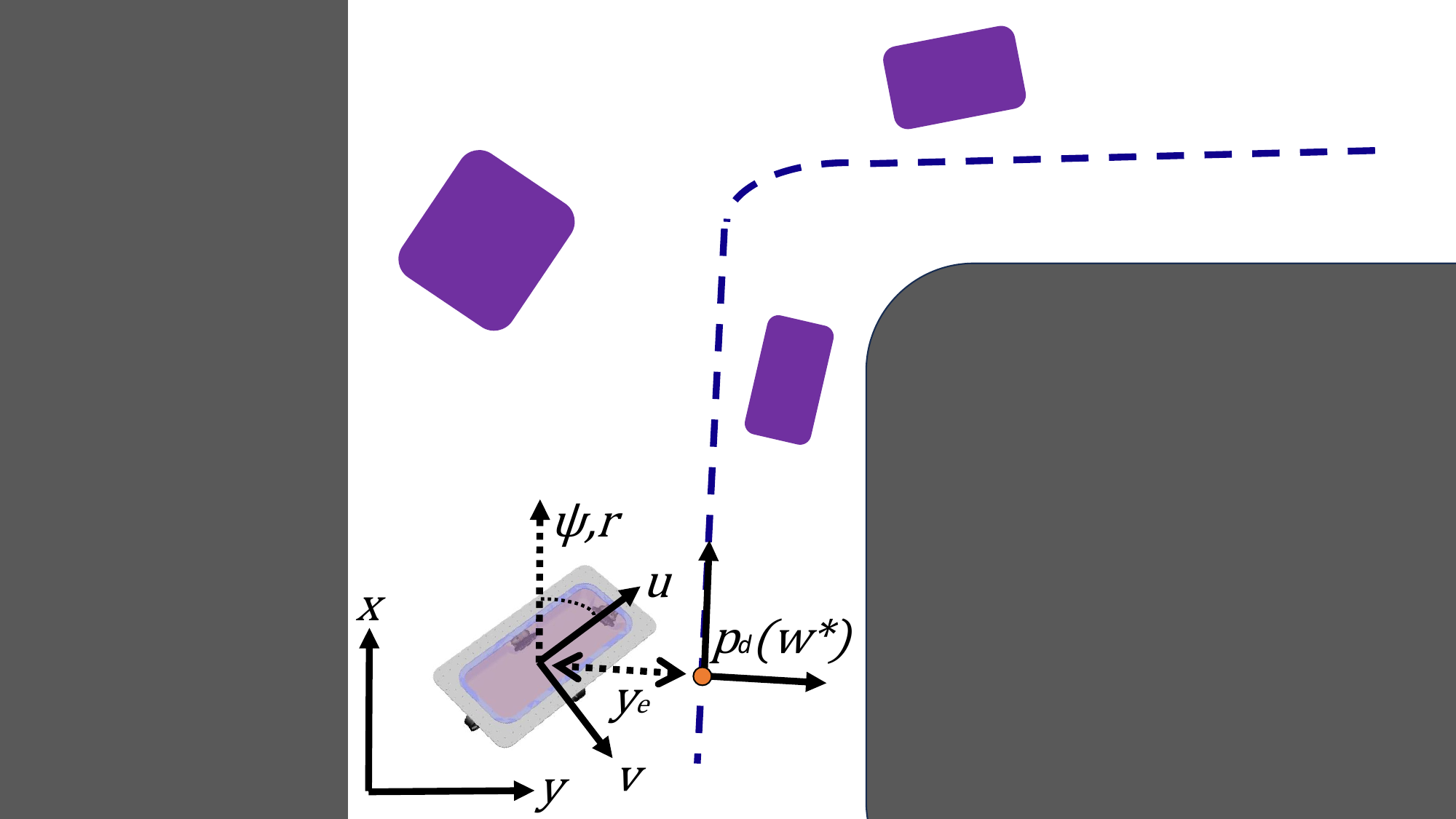}
    \caption{The ASV with known areas for the waterway (white) and land (gray), a reference path (blue), and unknown floating obstacles (purple).}
    \label{problem}
\end{figure}

The ASV operates in an environment where the boundaries of the waterway (canal, river, etc.) are assumed to be known. Still, unknown floating static or quasi-static obstacles have to be detected by local sensing (\textit{e.g.}, LiDAR). Thus, inequality constraints should be defined to represent collision avoidance constraints for each waterway boundary $j$ and for each detected obstacle $i$, as $b_{bj}\geq 0$ and $b_{bi}\geq 0$, respectively. The local planning problem is to compute a trajectory that can remain close to the original path but avoids collisions with the spatial boundaries and the floating obstacles. Fig.~\ref{problem} illustrates the problem.

\subsection{High Order Control Barrier Functions} 
In this subsection, we briefly introduce the concept of high-order CBFs. If interested, the reader is referred to \cite{xiao2021high} for definitions of relative degree, forward invariance, and class $\mathcal{K}$ functions. 

Consider an affine control system:
\begin{equation}
\dot{\mathbf{x}}=f(\mathbf{x})+g(\mathbf{x})\mathbf{u}, \label{eqn:affine}%
\end{equation}
where $\mathbf{x}\in X\subset\mathbb{R}^{n}$, $f:\mathbb{R}^{n}\rightarrow\mathbb{R}^{n}$ and $g:\mathbb{R}^{n}\rightarrow\mathbb{R}^{n\times q}$ are Lipschitz continuous, and $\mathbf{u}\in U\subset\mathbb{R}^{q}$ is the control constraint set.


For a constraint $b(\mathbf{x})\geq0$ with relative degree $m$ --- \textit{i.e.}, we need to differentiate $b(\bm x)$ $m$ times along the dynamics (\ref{eqn:affine}) until the control first shows up in the corresponding derivative --- $b:\mathbb{R}^{n}\rightarrow\mathbb{R}$, and $\psi_{0}(\mathbf{x}):=b(\mathbf{x})$, we first define a sequence of CBFs $\psi_{i}:\mathbb{R}^{n}\rightarrow\mathbb{R},i\in\{1,\dots,m\}$ in the form:
\begin{equation}
\begin{aligned} \psi_i(\mathbf{x}) := \dot \psi_{i-1}(\mathbf{x}) + \alpha_i(\psi_{i-1}(\mathbf{x})),i\in \{1,\dots,m\}, \end{aligned} \label{eqn:functions}%
\end{equation}
where $\alpha_{i}(\cdot),i\in\{1,\dots,m\}$ denotes a  class $\mathcal{K}$ function of $(m-i)^{th}$ order differentiable.
We then define a sequence of safe sets $C_{i}, i\in\{1,\dots,m\}$ corresponding to (\ref{eqn:functions}) in the form:
\begin{equation}
\label{eqn:sets}\begin{aligned} C_i := \{\mathbf{x} \in \mathbb{R}^n: \psi_{i-1}(\mathbf{x}) \geq 0\}, i\in\{1,\dots,m\}. \end{aligned}
\end{equation}

\begin{definition}
\label{def:hocbf} (\textit{High Order Control Barrier Function (HOCBF)}
\cite{xiao2021high}) Let $\psi_{i}(\mathbf{x}), i\in\{1,\dots, m\}$ be defined by \eqref{eqn:functions} and $C_{i}, i\in\{1,\dots, m\}$ be defined by \eqref{eqn:sets}. A function $b: \mathbb{R}^{n}\rightarrow\mathbb{R}$ is defined as a High Order Control Barrier Function (HOCBF) of relative degree $m$ for system (\ref{eqn:affine}) if there exist $(m-i)^{th}$ order differentiable class $\mathcal{K}$ functions $\alpha_{i},i\in\{1,\dots,m\}$ such that: 
\begin{equation}
\label{eqn:constraint}\begin{aligned} \sup_{\mathbf{u}\in U}[L_f\psi_{m-1}(\mathbf{x}) + L_g\psi_{m-1}(\bm x)\mathbf{u} + \alpha_m(\psi_{m-1}(\mathbf{x}))] \geq 0, \end{aligned}
\end{equation}
for all $\mathbf{x}\in \cap_{i=1}^m C_{i}$. In the above, the left part is equivalent to $\psi_m(\mathbf{x})$, and $L_{f}$, $L_{g}$ denote Lie derivatives along $f$ and along $g$, respectively.
\end{definition}

The HOCBF is a general form of a CBF with relative degree one \cite{Aaron2014}. In other words, setting $m=1$ reduces the HOCBF to the common CBF form: $L_fb(\mathbf{x}) + L_gb(\mathbf{x})\mathbf{u} + \alpha_1(b(\mathbf{x}))\geq 0.$

\begin{theorem}
\label{thm:hocbf} (\textit{Safety guarantees with HOCBFs}\cite{xiao2021high}) Given an HOCBF $b(\mathbf{x})$ as in Def. \ref{def:hocbf} with the associated sets $C_{i}, i\in\{1,\dots,m\}$ defined by \eqref{eqn:sets}, if $\mathbf{x}(0) \in \cap_{i=1}^mC_{i}$, then any Lipschitz continuous controller $\mathbf{u}(t)\in U$ that satisfies the HOCBF constraint in (\ref{eqn:constraint}), $\forall t\geq0$ renders $\cap_{i=1}^mC_{i}$ forward invariant for system (\ref{eqn:affine}), and we have that $b(\bm x(t))\geq 0, \forall t\geq 0$.
\end{theorem}

\section{Methodology}

In this section, the proposed methodology for safe motion planning is described. 

\subsection{Architecture Overview}

\begin{figure}[tb]
    \centering
    \includegraphics[width=0.5\linewidth] {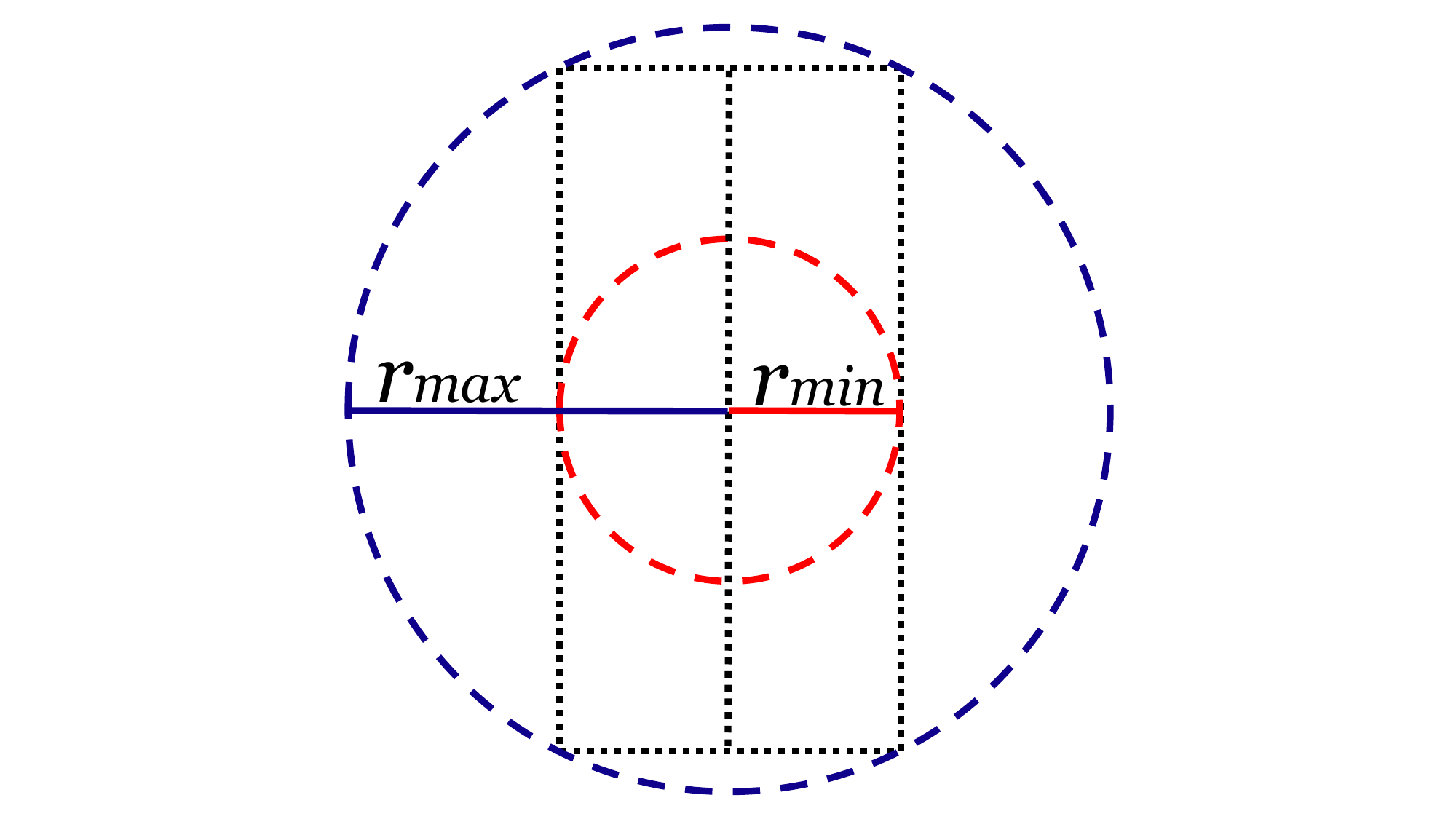}
    \caption{ASV footprint with enclosing radius $r_{\text{max}}$ (blue), and inner radius $r_\text{min}$ (red).}
    \label{circles}
\end{figure}

The proposed approach employs MPC to generate an initial trajectory while ensuring safety around floating obstacles using HOCBFs and a dynamic obstacle inflation methodology. First, unknown obstacles are detected using the LiDAR perception system, and ellipses are fitted around them. Then, an MPC problem is formulated, treating obstacles and waterway boundaries as constraints, with an inflation of the ASV's inner radius \( r_{\text{min}} \) (Fig.~\ref{circles}), defined as half the ASV's width plus a safety distance \( d_s \). The MPC provides an approximate initial trajectory that is not guaranteed to be safe or collision-free.

\begin{figure*}[tb]
    \centering
    \begin{subfigure}{0.28\textwidth}
        \includegraphics[width=\linewidth]{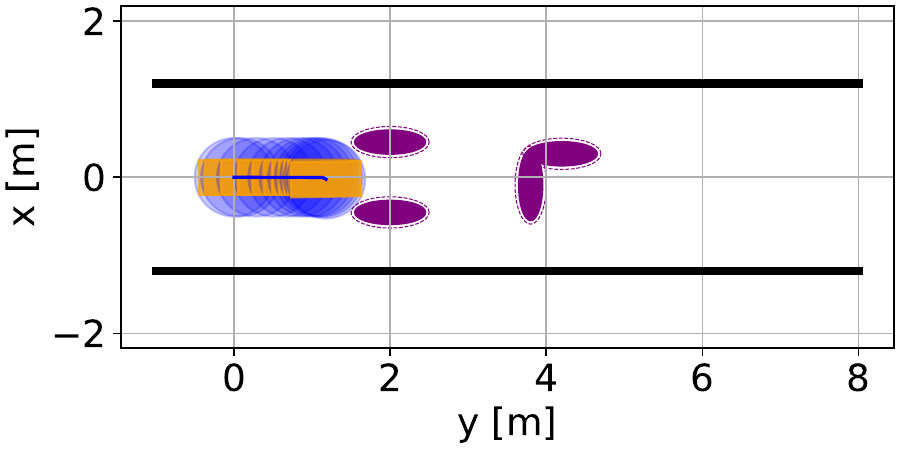}
        \caption{}
        \label{fig:sfirst}
    \end{subfigure}
    \hfill
    \begin{subfigure}{0.28\textwidth}
        \includegraphics[width=\linewidth]{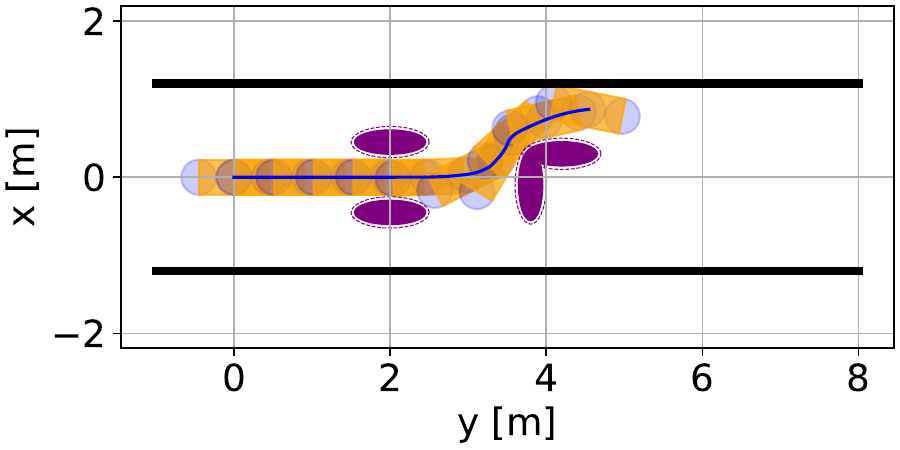}
        \caption{}
        \label{fig:ssecond}
    \end{subfigure}
    \hfill
    \begin{subfigure}{0.28\textwidth}
        \includegraphics[width=\linewidth]{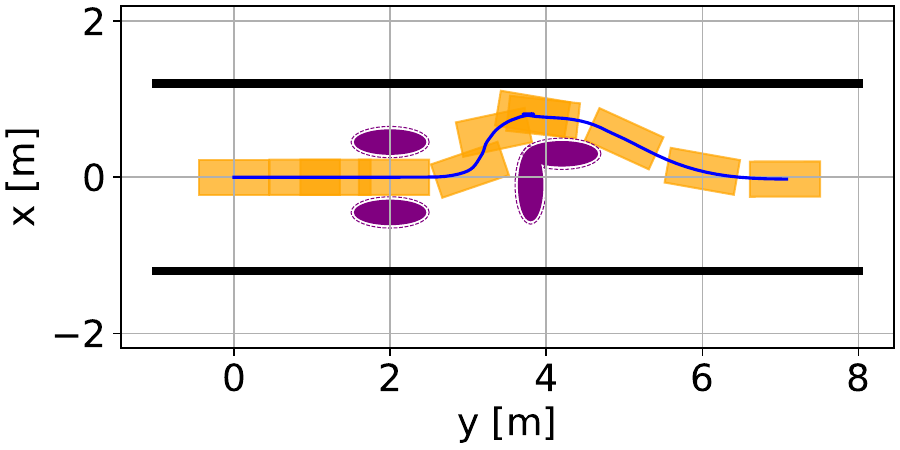}
        \caption{}
        \label{fig:sthird}
    \end{subfigure}
    \caption{Comparison between a) MPC under a fixed enclosing circle, b) MPC with a set of circles, c) the proposed MPC-CBF framework. Under narrow circumstances, the fixed inflation radius fails. The set of circles can solve for narrow corridors, but fails when more obstacles require more complex maneuverability. The proposed framework can solve for these cases.}
    \label{fig:comp}
\end{figure*}

Next, the proposed adaptive obstacle representation modifies the inflation radius, considering the relative position of each obstacle/boundary and the ASV (see implementation details in Section~\ref{detection}). After that, the HOCBF obstacle/boundary constraints are inflated by this new computed radius, which guarantees the vessel's safety, assuming it will not rotate on the spot. This assumption can lead to deadlock scenarios in tight spaces. Hence, a deadlock recovery algorithm is proposed to safely maneuver out of these narrow spaces (see Section~\ref{deadrec}). 
Algorithm~\ref{algorithm} illustrates the proposed framework at the end of this section.
    
\subsection{Model Predictive Control}
For the MPC formulation, the ASV dynamic model is first rearranged by merging (\ref{MFCPredynamicsA}) and (\ref{MFCPredynamicsB}), reformulating the dynamic model as \eqref{eqn:affine}, 
where $f(\mathbf{x})=[\bm{0}_{3\times 3}~\mathbf{R}(\bm{\eta});\bm{0}_{3\times 3}~ -\mathbf{M}^{-1}(\mathbf{C}(\mathbf{v})+\mathbf{D}(\mathbf{v}))]$, $g(\mathbf{x}) = [\bm{0}_{3\times 4};\mathbf{M}^{-1}\mathbf{B} ]$, the state vector $\mathbf{x}(t) = [x \quad y \quad \psi \quad  u \quad  v \quad  r]^{T} \in \mathbb{R}^{6}$ and the control vector $\mathbf{u}(t)=[f_1 \quad f_2 \quad f_3 \quad f_4]^{T}$. Rotated ellipse constraints describe the floating obstacles as:
\begin{equation} \label{ellipseconst}
    \frac{(x^\prime_i)^2}{a_{mi}^2} + \frac{(y^\prime_i)^2}{b_{mi}^2} \leq 1 ,
\end{equation}
Here, $x^\prime_i = (x - x_{oi})\cos\theta_{oi} + (y - y_{oi})\sin\theta_{oi}, y^\prime_i = (x - x_{oi})\sin\theta_{oi} - (y - y_{oi})\cos\theta_{oi}$, $x_{oi},y_{oi}$ is the position of the $i$-th obstacle, and $\theta_{oi}$ is its orientation. $a_{mi} = a_{oi} + r_{\text{min}}$ is the length of the inflated semi-major axis, $b_{mi} = b_{oi} + r_{\text{min}}$ is the length of the inflated semi-minor axis, and $a_{oi},b_{oi}$ are the lengths of the obstacle semi-major and semi-minor axes (the result of the obstacle fitter plus the safety distance $d_s$), respectively. The constraints for the waterway limits depend on the chosen spatial representation, as describing either a continuous tube as in \cite{Wei2024}, or rectangular corridors as in \cite{BOS20234877}. Then, a Nonlinear Program (NLP) is formulated as:
\begin{subequations} \label{OCP}
\begin{eqnarray}
    \begin{split}
        \min_{\substack{\mathbf{x}_0,\dots , \mathbf{x}_N,\\ \mathbf{u}_0,\dots , \mathbf{u}_{N-1}}} J(\mathbf{x},\mathbf{u}) = & \sum_{k=0}^{N-1} L(\mathbf{x}_k, \mathbf{u}_k) +  M(\mathbf{x}_N)
    \end{split} \\
\text{s.t.}~~\mathbf{x}_0= \hat{\mathbf{x}}_0,\\
\mathbf{x}_{k+1}=f(\mathbf{x}_k)+g(\mathbf{x}_k)\mathbf{u}_k,  k=0,\cdot \cdot \cdot, N-1,\\
\mathbf{x}_{k,\text{min}}\leq\mathbf{x}_k\leq \mathbf{x}_{k,\text{max}}, k=0,\cdot \cdot \cdot, N,\\
\mathbf{u}_{k,\text{min}}\leq\mathbf{u}_k\leq \mathbf{u}_{k,\text{max}}, k=0,\cdot \cdot \cdot, N-1, \\
\text{obstacle $i$ constraint \eqref{ellipseconst}}, k=0,\cdot \cdot \cdot, N, \\
\text{waterway boundary constraints}, k=0,\cdot \cdot \cdot, N.
\end{eqnarray}
\end{subequations}

Here, $\mathbf{x}_k \in \mathbb{R}^{n_q}$ denotes the vessel state,  $\mathbf{u}_k \in \mathbb{R}^{n_u}$ denotes the control input, $ \hat{\mathbf{x}}_0 \in \mathbb{R}^{n_q}$ denotes the current state estimate, $L(\mathbf{x}_k, \mathbf{u}_k)$ denotes the Lagrange objective term, and $ M(\mathbf{x}_N)$ the Mayer objective term. $L(\mathbf{x}_k, \mathbf{u}_k)$ is designed to take into account the path-following objectives of minimizing the cross-track error and the deviation from the angle path. Additionally, the cost function is extended to follow a desired speed $u_{\text{ref}}$, which can be a waterway limit or an operation preference, and to reduce rotational motion aggressiveness and control effort. Thus, it is defined as:
\begin{equation}
    \begin{split}        
    L(\mathbf{x}_k&, \mathbf{u}_k) = Q_{y}y_{e,k}^2 +Q_rr_k^2 + Q_\psi((\sin(\psi_k)-\sin(\gamma_{p,k}))^2 \\ &+ (\cos(\psi_k) \cos(\gamma_{p,k}))^2) + Q_u(u_k-u_{\text{ref}})^2 +||\mathbf{u}_k||_\mathbf{W}^2 ,
    \end{split}
\end{equation}
where $Q_{y},Q_\psi,Q_u,Q_r$ are penalty weights, and $\mathbf{W}$ is the control input weighting matrix. Similarly, the Mayer objective term is defined as:
\begin{equation}
    \begin{split}        
    M(\mathbf{x}_N) = &Q_{y}y_{e,N}^2 + Q_rr_N^2 + Q_\psi((\sin(\psi_N)-\sin(\gamma_{p,N}))^2 \\
    &+ (\cos(\psi_N) -\cos(\gamma_{p,N}))^2) + Q_u(u_N-u_{\text{ref}})^2 .
    \end{split}
\end{equation}
The MPC formulation was prototyped in Python using the \texttt{Rockit} toolbox \cite{gillis2020effortless} and the trajectory optimization solver \texttt{FATROP} \cite{vanroye2023fatrop}, with its C-code generation capability.

\subsection{Adaptive Obstacle Representation}

The proposed MPC formulation provides a solution that does not account for the complete ASV footprint, risking collisions. Nevertheless, a conventional solution would be conservative, inflating the obstacles using the enclosing circle with radius $r_\text{max}$ (Fig.~\ref{circles}). Therefore, a methodology is proposed to change the inflation radius depending on the relative position of the obstacle and the current ASV orientation. First, an assumption is established to leverage the overactuation of the vessel:
\begin{assump} \label{assumption}
    As the ASV approaches an obstacle initially, there will be no significant or aggressive rotations.
\end{assump}

Under this assumption, the ASV can be treated geometrically as a rectangle with a fixed heading angle. Then, the next step is to compute the closest point between the ASV and the obstacle ellipse, as this would be the first point to create a collision. Finding the closest points between a rectangle and an ellipse is an optimization problem that can be efficiently solved numerically using Newton's method or gradient descent. Once the closest points $[x^*,y^*]^T,[x^*_o,y^*_o]^T$ are found, the relative angle $\alpha_o$ between the ASV and the obstacle closest contact points is computed with $\phi_o = \mathrm{atan2}((y^*_o-y^*),(x^*_o-x^*))$ and $\alpha_o = \psi - \phi_o.$ 
Finally, the projection of the ASV width and length is used to compute the ASV span in the direction toward the obstacle, which divided by two gives the dynamic inflation radius $r_o(\alpha_o)$: 
\begin{equation} \label{dynamicradius}
    r_o(\alpha_o) = (w|\sin\alpha_o| + l|\cos\alpha_o|)/2.
\end{equation}

\begin{remark}
    The adaptive obstacle representation is designed around Assumption~\ref{assumption}, which limits the solution to fully-actuated systems (as many inland waterways vessels are \cite{Wei2023,ZHANG2024229}). The no-rotation assumption allows the vessel to move longitudinally and laterally to counteract obstacles and guarantee safety. The boat can control its orientation; thus, the assumption can become a system constraint. However, understanding that maneuverability is needed in some tight spaces, the deadlock recovery algorithm in Section~\ref{deadrec} is presented to perform safe rotations.
\end{remark}

\begin{figure}[tb]
    \centering
    \includegraphics[width=0.5\linewidth] {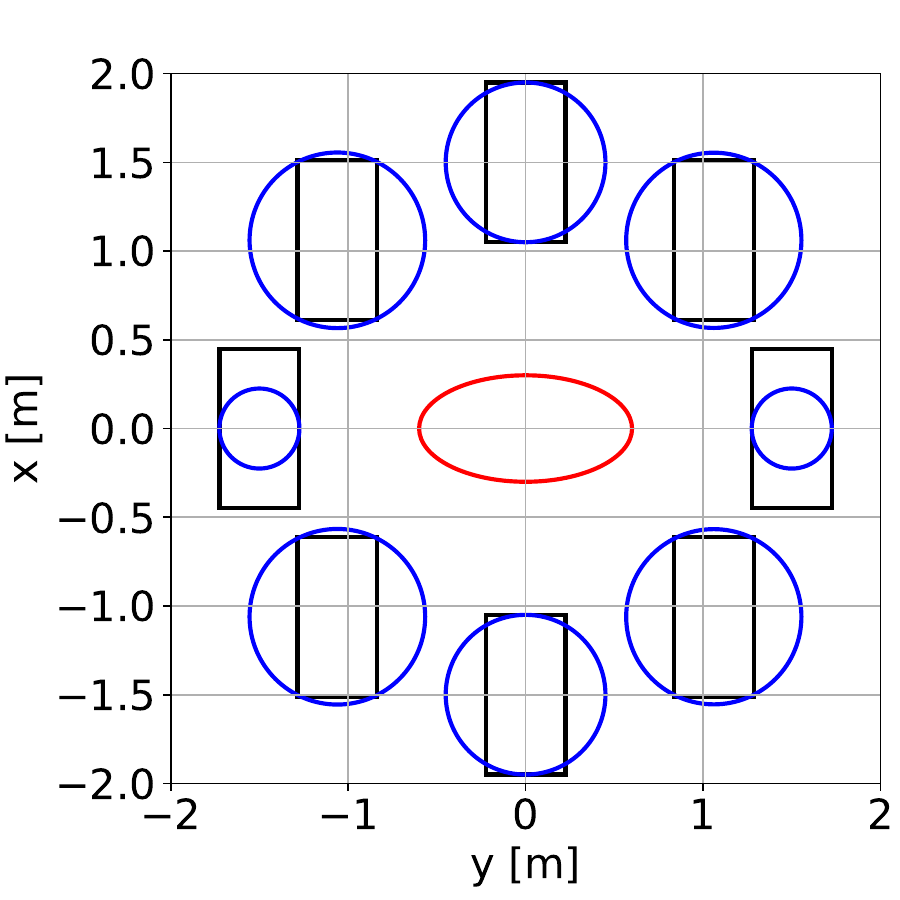}
    \caption{Examples of the dynamic radius inflation. The ASV footprint is shown in black, the red ellipse represents the obstacle plus a safety distance, and the blue circles represent the considered inflation.}
    \label{inflation}
\end{figure}

\begin{remark}    
The adaptive inflation \eqref{dynamicradius} establishes that if the ASV is facing directly towards an obstacle, then the distance between the center of the vessel and the obstacle circumference should be at least half the vessel length. If the obstacle is at a 90-degree angle with respect to the boat, then the obstacle should be at a distance of at least half the ASV's width. This equation modifies the constraint dimension such that the ASV can navigate through narrow corridors, in contrast to overly-conservative representations (see Fig.~\ref{fig:comp}). Fig.~\ref{inflation} illustrates the equation.
\end{remark}

\subsection{High Order Control Barrier Functions}

Since the MPC provides an approximate solution with no collision-free guarantees for the full dimension of the ASV, HOCBFs are proposed to ensure safety. Following the MPC constraint \eqref{ellipseconst}, a barrier function $b_{oi}(\mathbf{x})$ is designed for each obstacle $i$ as an ellipse:
\begin{equation}\label{eqn:safe1}
    b_{oi}(\mathbf{x}) = \frac{(x^\prime_i)^2}{a_{bi}^2} + \frac{(y^\prime_i)^2}{b_{bi}^2} - 1,
\end{equation}
where $a_{mi} = a_{oi} + r_{oi}(\alpha_{oi}), b_{mi} = b_{oi} + r_{oi}(\alpha_{oi})$ are the length of the inflated semi-major and semi-minor axes, respectively. In the case of waterway boundaries, similar to the MPC constraints, they can be defined in terms of tubes or corridors. For demonstration purposes, this work considers scenarios that simulate straight, narrow canals. Thus, the system should satisfy the constraints $-(h_c)/2 + r_{ot}(\alpha_{ot}) \leq x + x_c \leq (h_c)/2-r_{ob}(\alpha_{ob})$, $-(w_c)/2 + r_{ol}(\alpha_{ol}) \leq y + y_c \leq (w_c)/2-r_{or}(\alpha_{or})$, where $(x_c,y_c)$ is the center of the canal, $h_c,w_c$ are the canal height and width, and $r_{oj}(\alpha_{oj}),r_{oj}(\alpha_{oj})$, $\forall j = t,b,l,r$ are computed with $\phi_{oj}=\{0, \pi, -\pi/2, \pi/2\}$ rad, respectively. Then, four barrier functions are defined as:
\begin{equation}\label{eqn:safe2}
    \begin{aligned}
        b_{bb} &= (h_c)/2 - r_{ob}(\alpha_{ob}) + \Delta_x, & b_{bt} &= (h_c)/2 - r_{ot}(\alpha_{ot}) - \Delta_x, \\
        b_{bl} &= (w_c)/2 - r_{ol}(\alpha_{ol}) + \Delta_y, & b_{br} &= (w_c)/2 - r_{or}(\alpha_{or}) - \Delta_y.
    \end{aligned}
\end{equation}
where $\Delta_x = x + x_c, \Delta_y = y + y_c$. The relative degrees of the safety constraints (\ref{eqn:safe1})-(\ref{eqn:safe2}) are all 2, \textit{i.e.}, $m = 2$ in the HOCBF (\ref{eqn:constraint}).
The corresponding HOCBF constraints can then be defined as:
\begin{subequations} \label{eqn:hocbfs}
\begin{eqnarray}
L_f^{m}b_{oi}(\mathbf{x}) + L_gL_f^{m-1}b_{oi}(\mathbf{x})\mathbf{u} + O(b_{oi}(\mathbf{x})) \\ + \alpha_m(\psi_{bi,m-1}(\mathbf{x})) \geq 0, \nonumber \\
L_f^{m}b_{bj}(\mathbf{x}) + L_gL_f^{m-1}b_{bj}(\mathbf{x})\mathbf{u} + O(b_{bj}(\mathbf{x})) \\+ \alpha_m(\psi_{2,m-1}(\mathbf{x})) \geq 0, \nonumber
\end{eqnarray} 
\end{subequations}
where $\psi_{k, m-1}, k\in\{oi,bj\}$ are defined as in \eqref{eqn:functions} for $b_{oi}(\mathbf{x})$ and $b_{bj}(\mathbf{x})$, respectively. Any control input applied to the ASV should satisfy \eqref{eqn:hocbfs}. At each time step, the control input solution from the MPC is used as reference $\mathbf{u}_R$, for the following optimization:
\begin{equation} \label{qp}
	\begin{aligned}
	&\min_{\mathbf{u}} ||\mathbf{u} - \mathbf{u}_R||^2,
	&\text{ s.t., (\ref{eqn:hocbfs})}.
	\end{aligned}
	\end{equation}
Hence, the HOCBFs compute a new control input $\mathbf{u}$. The optimization problem \eqref{qp} is a QP problem, which was implemented in this work in C++ using the software package \texttt{qpOASES} \cite{Ferreau2014}.

\subsection{Deadlock Recovery Algorithm} \label{deadrec}

The proposed methodology takes advantage of the capability of ASV holonomic motion. However, not utilizing rotations to avoid obstacles may lead to deadlock situations. Henceforth, a deadlock recovery algorithm is introduced. First, a stuck-detection algorithm is used. This algorithm keeps track of a number of the vessel's previous positions and calculates the average of such positions. If the maximum distance between the average and the rest of the positions is lower than a threshold, then the ASV is assumed to be stuck, and a flag is raised that the vessel is in a narrow passage. This heuristic should be tuned according to the expected motion of the ASV. The deadlock recovery maneuver is activated if the vessel is inside a narrow passage. The maneuver first computes an approximate angle that may liberate the boat. This computation assumes that the ASV can only be stuck in two scenarios: 1) between two obstacles, or 2) between an obstacle and one of the waterway boundaries. Thus, the stuck scenario is assessed using the closest points between the ASV to each obstacle and each waterway boundary. According to the scenario, the two closest points $\pmb{p_{s1},p_{s2}}$ between the closest elliptical obstacle and the second object (second closest obstacle or closest waterway boundary) are computed, \textit{i.e.}, the nearest points between two ellipses or between an ellipse and a straight line. Next, the normal of the angle between them is computed:
\begin{subequations} \label{newangle}
\begin{eqnarray}
        &&\phi_s = \mathrm{atan2}((y_{s1}-y_{s2}),(x_{s1}-x_{s2})), \\
        &&\phi_+ = \phi_s + \pi/2 - \psi, \\
        &&\phi_- = \phi_s - \pi/2 - \psi, \\
        && \psi_d = \begin{cases}
            \phi_s + \pi/2, & \text{if} |\phi_+| \geq |\phi_-|,  \\
            \phi_s - \pi/2, & \text{if} |\phi_+| < |\phi_-|,
        \end{cases}
\end{eqnarray}
\end{subequations}
and a Control Lyapunov Function (CLF), used in conjunction with CBFs \cite{Aaron2014} \cite{xiao2021high}, is defined by choosing $V = (r + k_v(\psi - \psi_d))^2$, with parameter $k_v>0$, resulting in a modification of \eqref{qp} as:
\begin{subequations} \label{qp2}
\begin{eqnarray}
    &&\min_{\mathbf{u}} ||\mathbf{u} - \mathbf{u}_R||^2 \\
	&&\text{ s.t. (\ref{eqn:hocbfs})}, \\
    &&~~~~~ L_vV(\mathbf{x}) + L_gV(\mathbf{x})\mathbf{u} + c_3V(\mathbf{x}) \leq 0.
\end{eqnarray}
\end{subequations}
Finally, by verifying the distance between the vessel and the objects during the recovery procedure, the ASV is monitored to determine if it is still in the narrow passage or if it has managed to navigate it. 

\begin{algorithm}[]
\begin{algorithmic}[1]
    \State Project LiDAR points into a short-term memory grid.
    \State Fit ellipses on clustered points.
    \State Solve an MPC problem formulation with the ASV inner radius inflation \eqref{OCP}.
    \State Compute the non-conservative radius for each obstacle $i$ with \eqref{dynamicradius}.
    \State Solve the Quadratic Programming (QP) problem \eqref{qp} with the updated HOCBF constraints, using the MPC solution as reference.
    \State Determine if the robot is stuck.
    \If { Robot is stuck}
        \State Compute approximate heading for deadlock recovery with \eqref{newangle}.
        \State Raise the flag that the ASV is in a narrow passage.
    \EndIf
    \While{ ASV in a narrow passage }
        \State Check if the robot is out of the narrow passage.        
        \State Perform steps 1-4, and solve the QP problem \eqref{qp2} to rotate the vessel to the computed desired heading safely.
    \EndWhile
  \end{algorithmic}
  \caption{Proposed safe motion planning algorithm.}
  \label{algorithm}
\end{algorithm}

\section{Results}

In this section, the simulation and experimental setup are described, and the results are discussed.


\subsection{Setup}

\subsubsection{Simulation Setup}
The simulation environment uses ROS, Rviz, and Gazebo to simulate LiDAR sensor information and for visualization, whereas the dynamic model \eqref{MFCPredynamicsA}-\eqref{MFCPredynamicsB} is programmed into a C++ ROS node. A Gazebo world was built with a lake visualization, representing an intersection scenario with canal boundaries. Different environments were designed considering various obstacle types such as boxes, poles, and buoys. The simulation environment was designed to use the same ROS codebase architecture as with the experimental platform. Simulations were carried out using an Intel(R) Core(TM) i7-12800H $@$ 4.8GHz, running the control loop at a frequency of 10 Hz,

\subsubsection{Experimental Setup}
\begin{figure}[tb]
    \centering
    \includegraphics[width=0.8\linewidth,trim={150 300 300 50},clip]{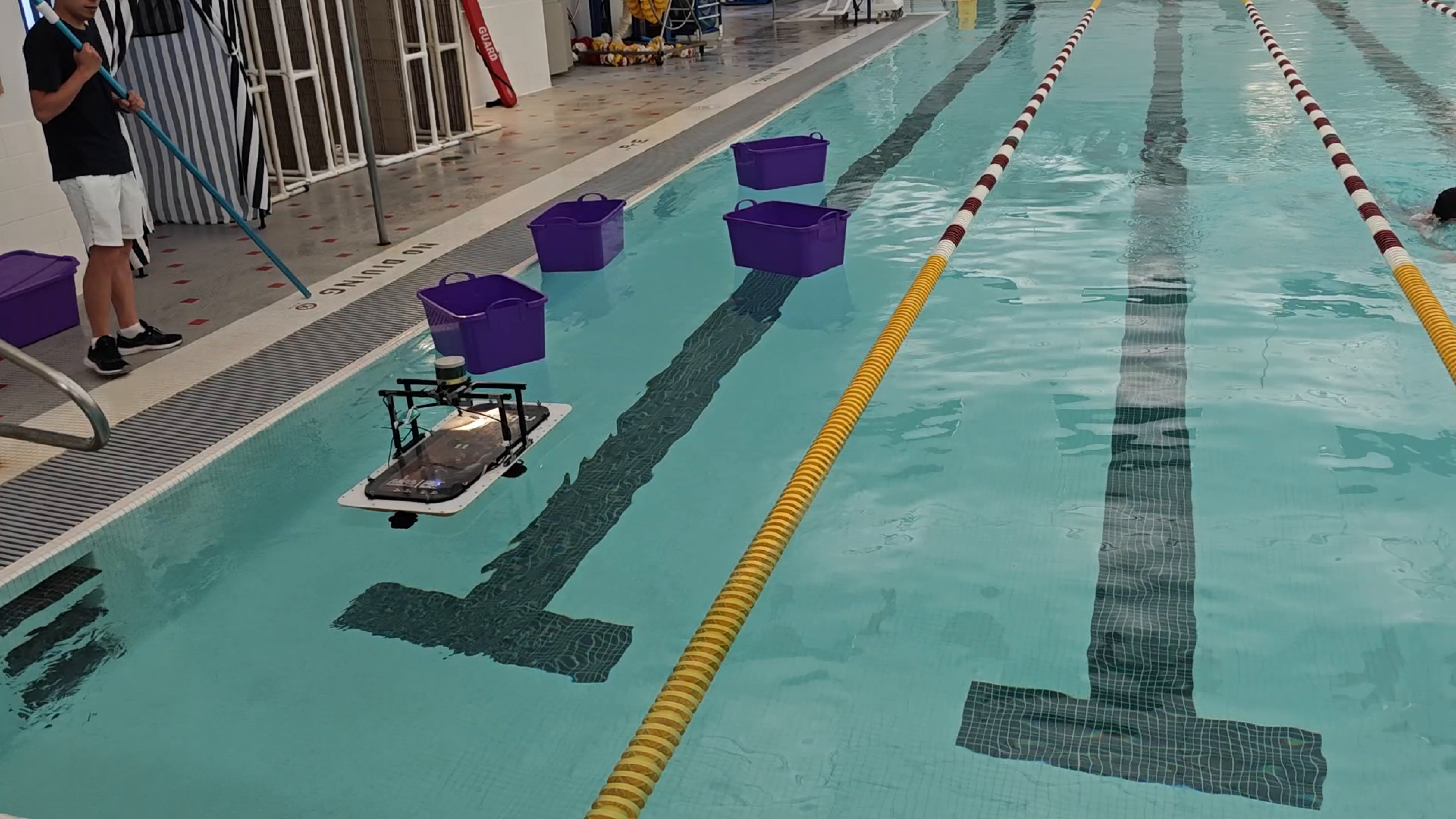}
    \caption{Experimental setup illustration.}
    \label{fig:expsetup}
\end{figure}

The Quarterscale Roboat \cite{WeiICRA2018} was deployed on a swimming pool. A swimming lane with a width of 2.5 m was selected to test, mimicking a narrow straight waterway. Four plastic boxes were placed inside the lane, creating narrow spaces between the boxes, and between boxes and the lane boundaries, as seen in Fig.~\ref{fig:expsetup}. 
In addition, the boxes were not anchored, which formed a quasi-static but dynamic environment. Localization is achieved using a LiDAR-Inertial-Odometry algorithm \cite{TixiaoIROS2020a}.

\subsubsection{Obstacle Detection System} \label{detection}
An online obstacle detection system using LiDAR data was implemented to test the motion planning algorithm in simulations and experiments, based on the perception system used in \cite{shan2020}. The LiDAR data is filtered within a desired $xyz$ range, and then projected into a 2D grid. The grid map is then inflated by $r_\text{min}+d_s$, creating clusters of what could represent a single obstacle (as no assumptions about the shape or size of the obstacles are made). Next, ellipses are fitted around contours using OpenCV \cite{opencv_library}. Finally, the ellipses are deflated by $r_\text{min}$, resulting in what is assumed to be each ellipse that contains a detected obstacle plus a safety distance.

\subsection{Simulation Results}

\begin{figure*}[!htb]
    \centering
    \begin{subfigure}{0.22\textwidth}
        \includegraphics[width=\linewidth]{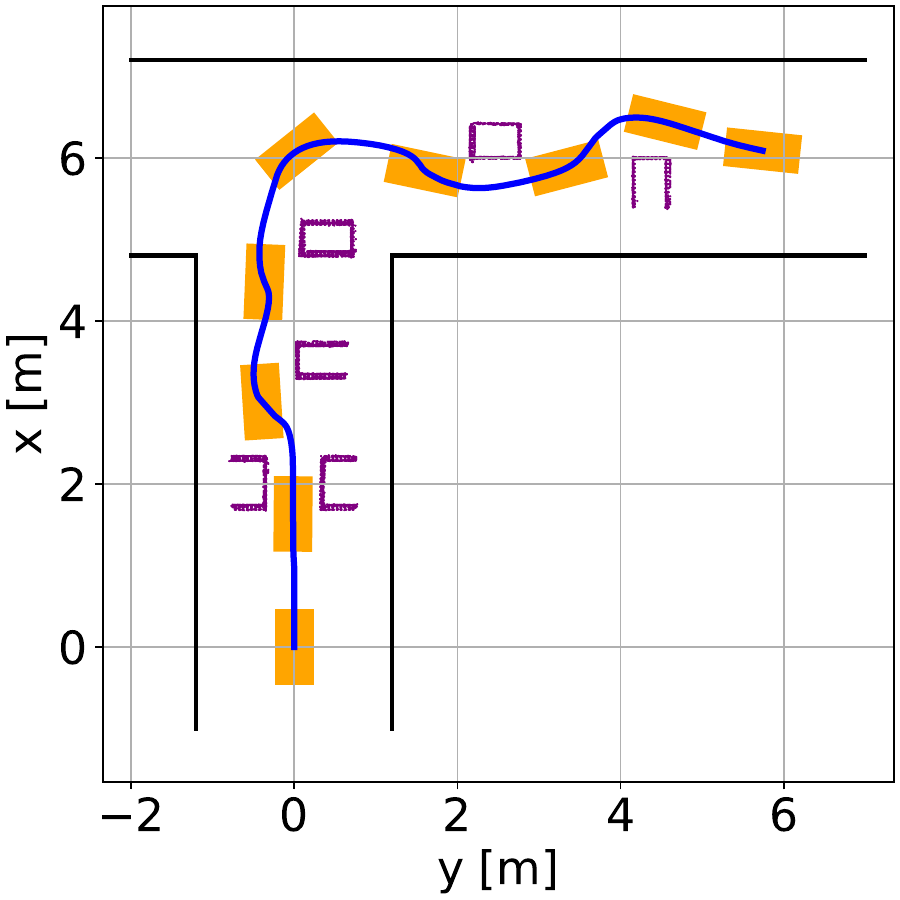}
        \caption{}
        \label{fig:sfirst}
    \end{subfigure}
    \hfill
    \begin{subfigure}{0.22\textwidth}
        \includegraphics[width=\linewidth]{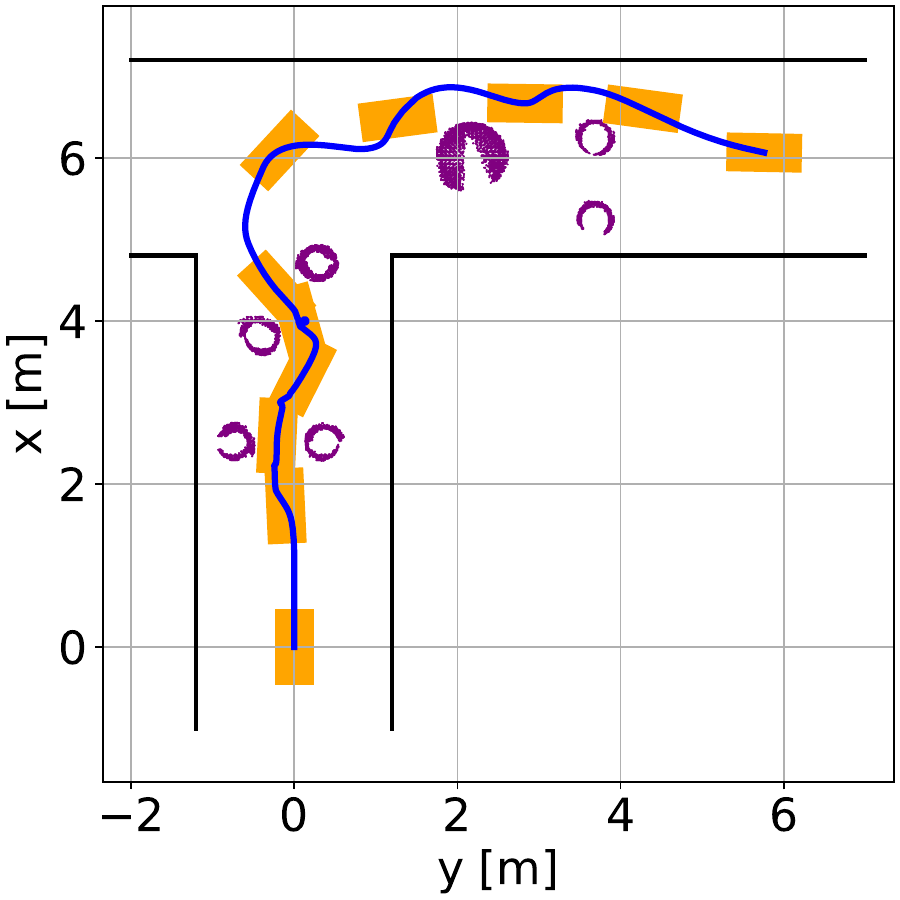}
        \caption{}
        \label{fig:ssecond}
    \end{subfigure}
    \hfill
    \begin{subfigure}{0.22\textwidth}
        \includegraphics[width=\linewidth]{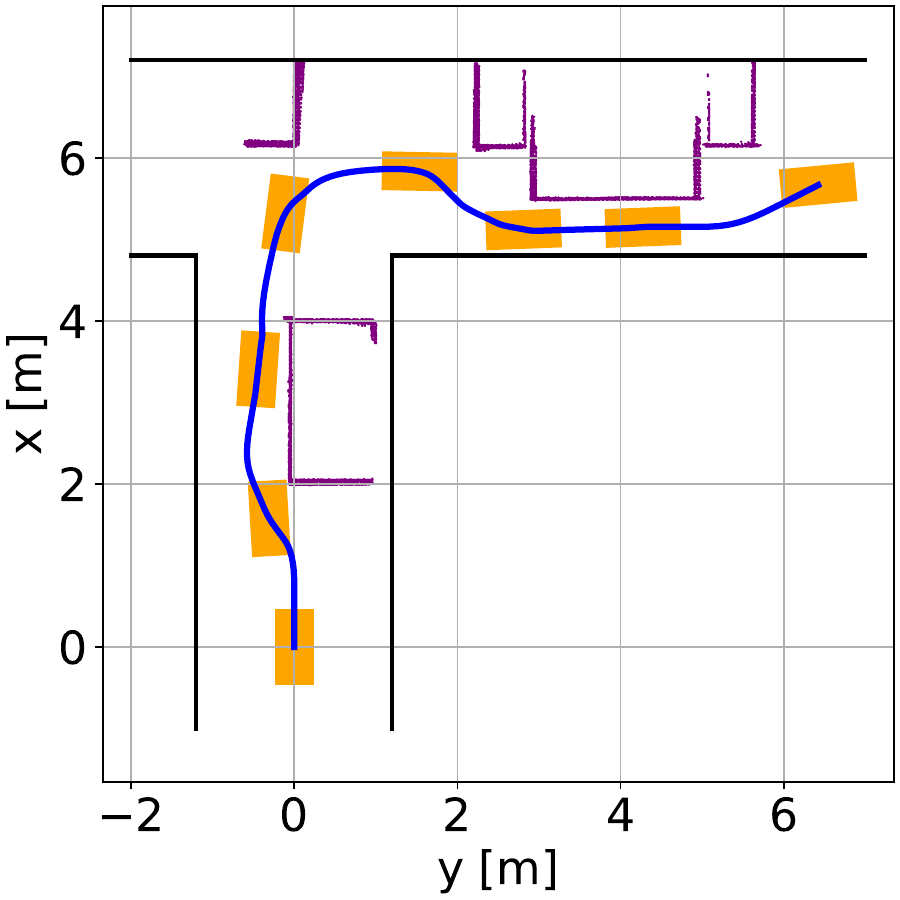}
        \caption{}
        \label{fig:sthird}
    \end{subfigure}
    \caption{Simulation results.}
    \label{simulationres}
\end{figure*}

Through all simulation runs, the vessel was tasked to follow a curve starting at coordinates $(0,0)$ and ending at $(6,6)$. In all environments, obstacles are positioned to create narrow spaces between them and between obstacles and the canal boundaries. Fig.~\ref{simulationres} shows the results on three different obstacle configurations. 
Here, the ASV is represented as the orange rectangle, the LiDAR point cloud is shown in purple, and the trajectory is given in blue. For all scenarios, the ASV maneuvered the obstacle courses without collisions. 
Table~\ref{tab1} includes computational performance data.

\begin{table*}[tb]
    \centering
    \resizebox{\textwidth}{!}{%
    \begin{tabular}{|l|c|c|c|c|c|c|}
    \hline
    \textbf{Computation} & \textbf{Fixed MPC} & \textbf{Multi. Circles} & \textbf{MPC-CBF} & \textbf{HOCBF QP} & \textbf{Closest points} & \textbf{Closest points (per obs)} \\
    \hline
    \textbf{Median (ms)} & 2.281 & 8.284 & 18.187 & 0.054 & 15.852 & 3.963 \\
    \textbf{Max (ms)} & 26.601 & 66.794 & 39.008 & 0.18 & 12.227 (max case) & 3.056 (max case) \\
    \hline
    \end{tabular} 
    }
    \caption{Solution times for different computations on simulation, including: MPC with a fixed enclosing circle; MPC with a set of circles; full MPC-CBF framework; adaptive HOCBF QP; computation of the closest points between the ASV and all obstacles; average computation of the closest points between the ASV and one obstacle. }\label{tab1}
\end{table*}

\subsection{Experimental Results}

\begin{figure}[!htb]
    \centering
    \begin{subfigure}{\linewidth}
        \centering
        \includegraphics[width=0.95\linewidth]{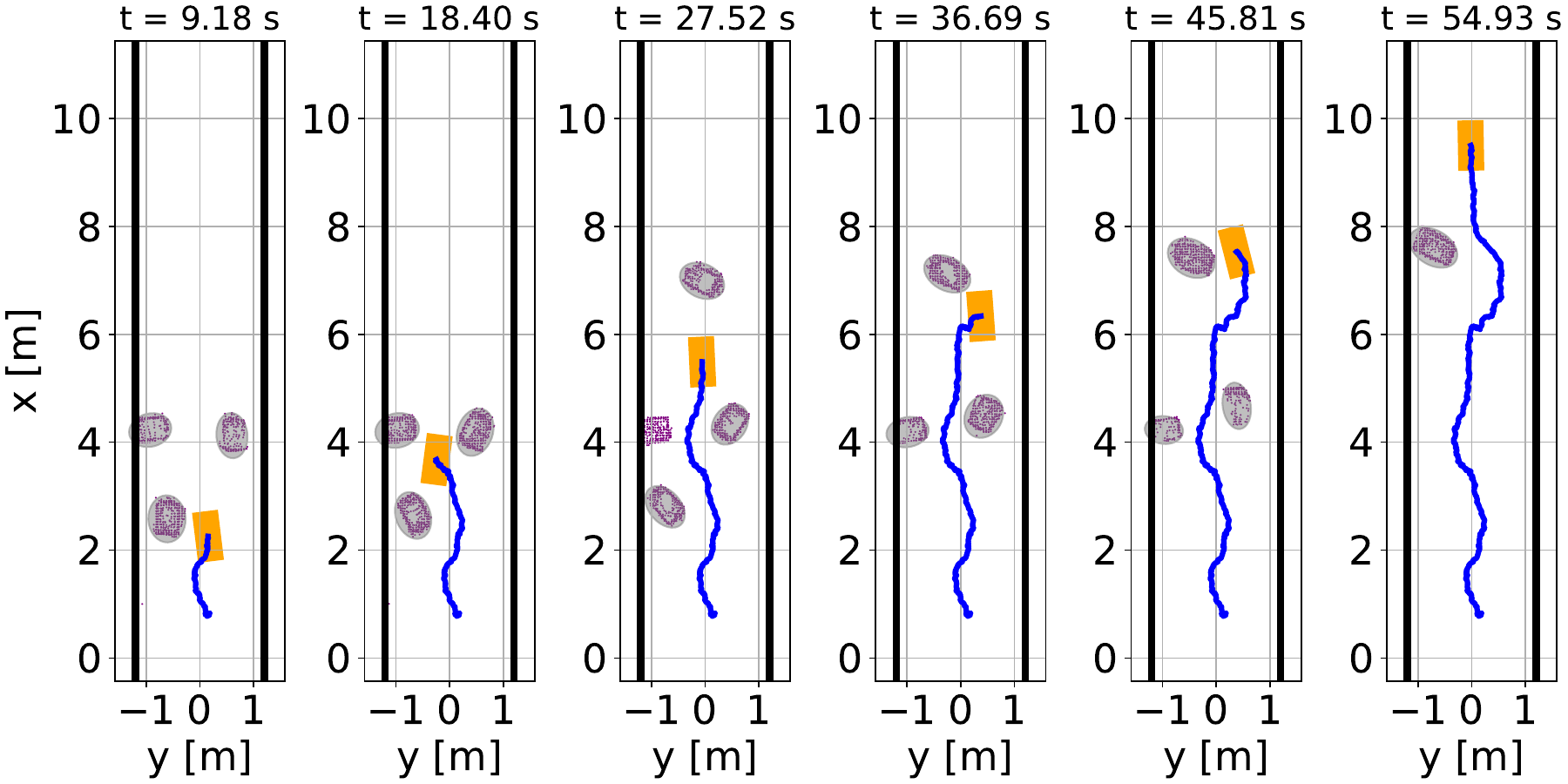}
        \caption{}
        \label{fig:efirst}
    \end{subfigure}
    \hfill
    \begin{subfigure}{\linewidth}
        \centering
        \includegraphics[width=0.95\linewidth]{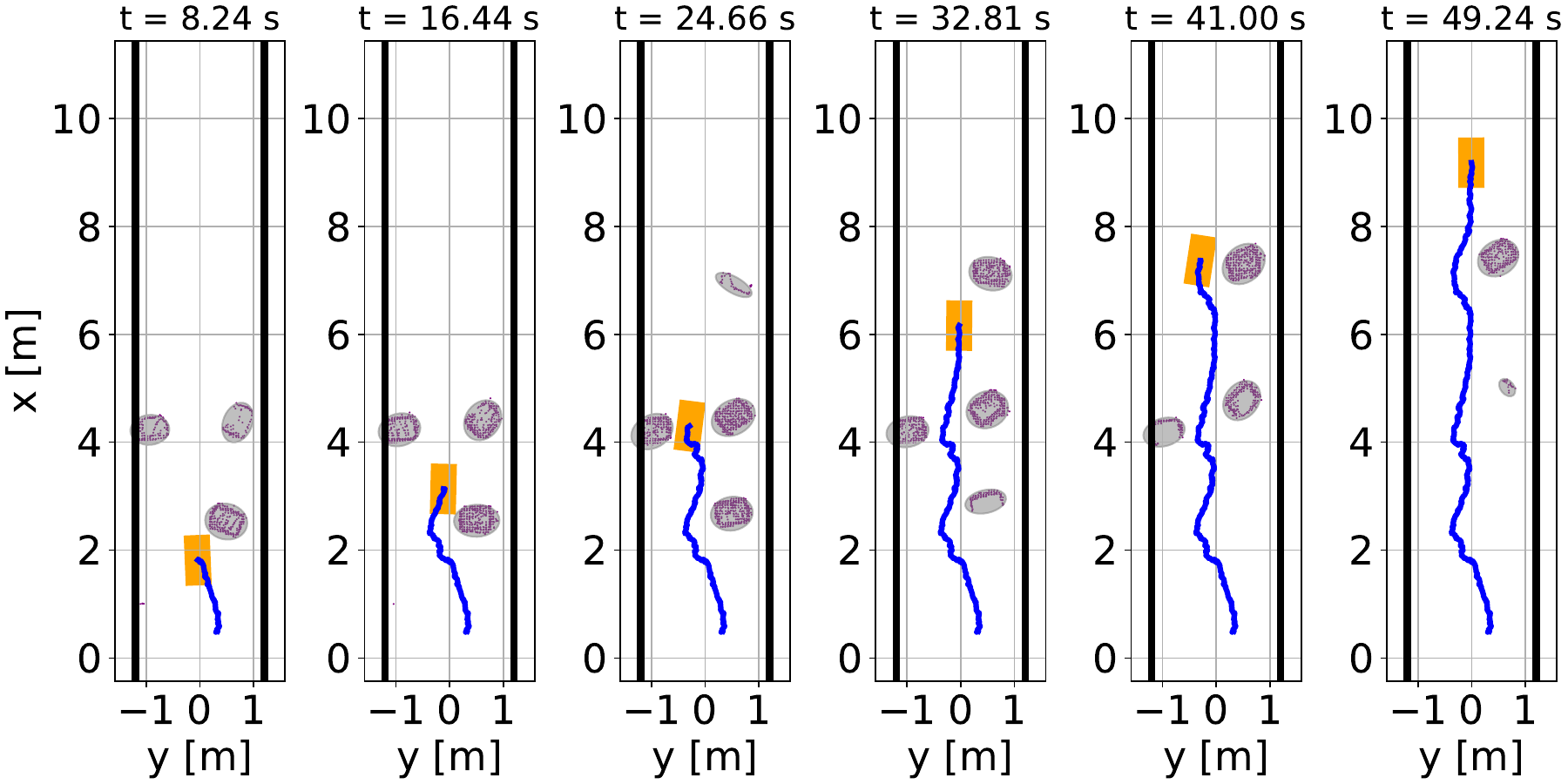}
        \caption{}
        \label{fig:esecond}
    \end{subfigure}
    \begin{subfigure}{\linewidth}
        \centering
        \includegraphics[width=0.95\linewidth]{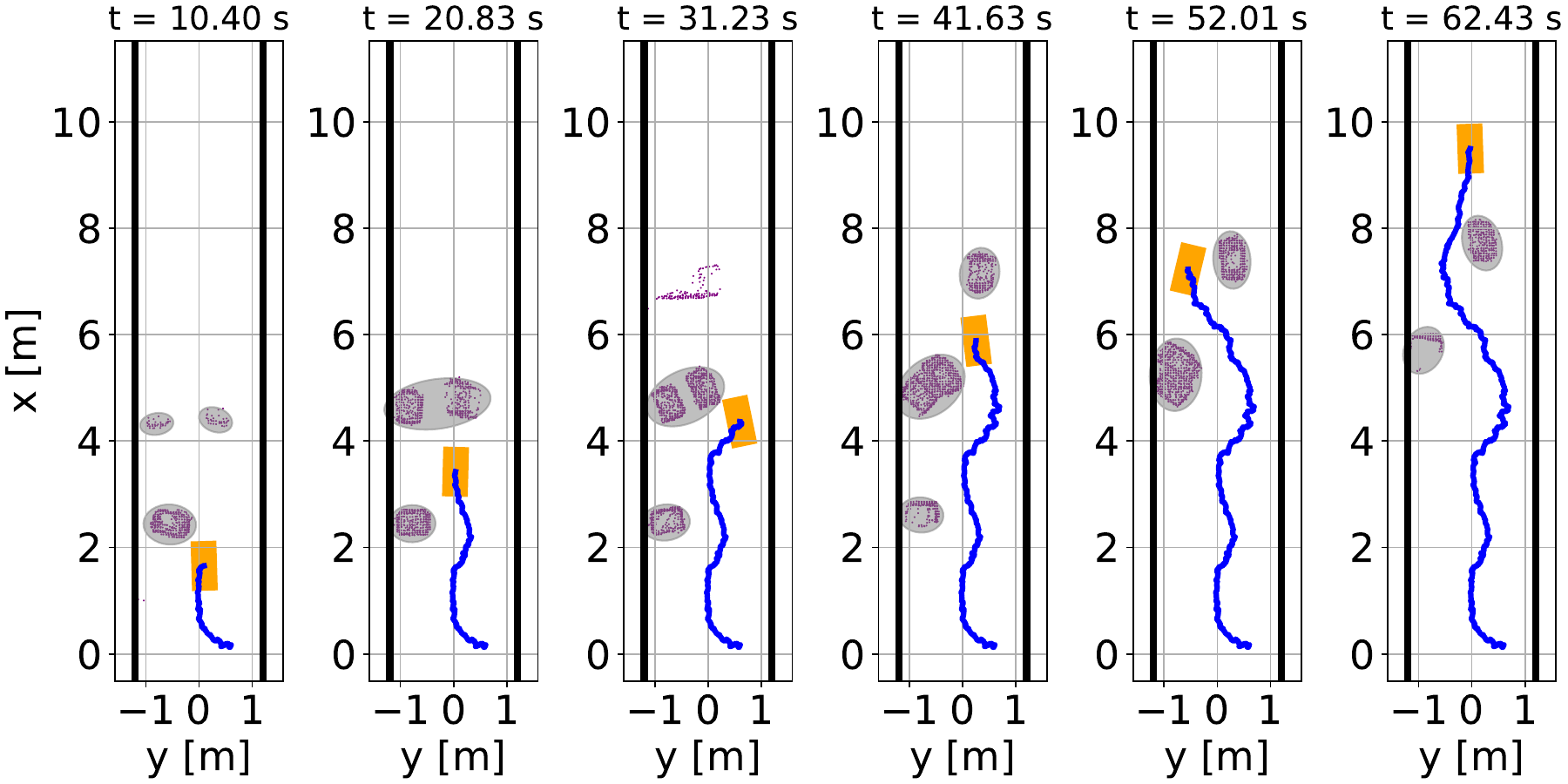}
        \caption{}
        \label{fig:ethird}
    \end{subfigure}
    \caption{Experimental results.}
    \label{experimentalres}
\end{figure}

Physical experiments were conducted in a swimming pool lane. The Roboat was tasked to follow a straight-line path through the center of the lane. Four plastic boxes were randomly positioned at each experimental run to obstruct the path. As the boxes were not anchored, the experiments showcase the performance of the proposed approach in dynamic environments. The experiment was repeated 10 times. Fig.~\ref{experimentalres} depicts three different experimental runs.
In addition to the ASV, its trajectory, and the LiDAR point cloud data, gray ellipses show the obstacle detection result at the given time step.

\section{Discussion}

The simulation results show the diverse sets of obstacles and scenarios that the proposed method can resolve without collision. In Fig.~\ref{simulationres}a-b, the closest obstacles to the origin act as a narrow gate, where the vessel can only navigate through by driving straight. In Fig.~\ref{simulationres}c, the large boxes simulate docks, containers, or docked vessels, which may obstruct large canal parts. 
Observing the solution times in Table~\ref{tab1}, the proposed approach implementation is capable of meeting real-time requirements (10-20Hz). However, the closest-points computation is the most expensive step, which could be implemented in parallel and further decrease the computation times. In this sense, solving the fixed-circle MPC and the QP together takes on average 2.281 ms. Moreover, the MPC with a set of circles can require longer computations, while still failing to solve for different tight scenarios, as in Fig.~\ref{fig:comp}.
Furthermore, although the method was designed for static obstacles, the unstructured nature of aquatic environments extends the results to quasi-static obstacles. The obstacles in the pool were slowly moving (speeds below 0.5 m/s). Still, the active perception system and the reactive nature of the proposed adaptive HOCBFs allowed the system to navigate without collisions. We report a 100\% success rate out of 10 physical experiments. However, it should be clarified that there were instances in which the ASV got stuck for a few seconds due to insufficient space to pass through. However, the natural motion of the obstacles eventually created feasible gaps for the vessel to navigate in between. As seen in Fig.~\ref{experimentalres}, the ASV consistently drove through narrow spaces between obstacles or between an obstacle and a lane. The accompanying video shows numerous successful experimental runs. Finally, although the method reduces conservativeness in contrast to representations with a singular large circle or a set of circles, there is still reliance on ellipsoidal shapes to represent obstacles. Hence, the practical performance of the proposed strategy depends on the obstacle detection system. If the obstacle ellipsoid covers much free space, this method will not guarantee solving for tight spaces. Nonetheless, if the perception system can provide solutions tailored to avoid the loss of free space, then the method has a higher chance of success. For instance, the method performs better if long rectangles are not treated as a single ellipsoid, but if each line within the field of view is represented as an independent ellipsoid.

\section{Conclusion}

In this paper, we propose a safe motion planning strategy for fully actuated ASVs that integrates Model Predictive Control (MPC) with Control Barrier Functions (CBFs). Our approach introduces an adaptive obstacle-representation methodology that retains the efficiency of ellipsoidal representations while reducing the conservativeness of traditional formulations. The proposed strategy was validated through both real-time simulations and real-world experiments. The results demonstrate that our method effectively navigates ASVs through narrow spaces and resolves deadlock situations in real-time, ensuring safe and reliable operation.

\bibliographystyle{IEEEtran}

\bibliography{References}

\begin{thebibliography}{10}
\providecommand{\url}[1]{#1}
\csname url@samestyle\endcsname
\providecommand{\newblock}{\relax}
\providecommand{\bibinfo}[2]{#2}
\providecommand{\BIBentrySTDinterwordspacing}{\spaceskip=0pt\relax}
\providecommand{\BIBentryALTinterwordstretchfactor}{4}
\providecommand{\BIBentryALTinterwordspacing}{\spaceskip=\fontdimen2\font plus
\BIBentryALTinterwordstretchfactor\fontdimen3\font minus
  \fontdimen4\font\relax}
\providecommand{\BIBforeignlanguage}[2]{{%
\expandafter\ifx\csname l@#1\endcsname\relax
\typeout{** WARNING: IEEEtran.bst: No hyphenation pattern has been}%
\typeout{** loaded for the language `#1'. Using the pattern for}%
\typeout{** the default language instead.}%
\else
\language=\csname l@#1\endcsname
\fi
#2}}
\providecommand{\BIBdecl}{\relax}
\BIBdecl

\bibitem{vagale2021}
A.~Vagale, R.~Oucheikh, R.~T. Bye, O.~L. Osen, and T.~I. Fossen, ``Path
  planning and collision avoidance for autonomous surface vehicles i: a
  review,'' \emph{Journal of Marine Science and Technology}, 2021.

\bibitem{Domenighini2024}
C.~Domenighini, ``Autonomous inland navigation: a literature review and
  extracontractual liability issues,'' \emph{Journal of Shipping and Trade},
  2024.

\bibitem{Wei2023}
W.~Wang, D.~Fernández-Gutiérrez, R.~Doornbusch, J.~Jordan, T.~Shan, P.~Leoni,
  N.~Hagemann, J.~K. Schiphorst, F.~Duarte, C.~Ratti, and D.~Rus, ``Roboat
  {III}: An autonomous surface vessel for urban transportation,'' \emph{Journal
  of Field Robotics}, vol.~40, no.~8, pp. 1996--2009, 2023.

\bibitem{Cheng2021}
Y.~Cheng, M.~Jiang, J.~Zhu, and Y.~Liu, ``Are we ready for unmanned surface
  vehicles in inland waterways? the usvinland multisensor dataset and
  benchmark,'' \emph{IEEE Robotics and Automation Letters}, vol.~6, no.~2, pp.
  3964--3970, 2021.

\bibitem{shan2020}
T.~Shan, W.~Wang, B.~Englot, C.~Ratti, and D.~Rus, ``A receding horizon
  multi-objective planner for autonomous surface vehicles in urban waterways,''
  in \emph{2020 59th IEEE Conference on Decision and Control (CDC)}, 2020, pp.
  4085--4092.

\bibitem{Streichenberg2023}
L.~Streichenberg, E.~Trevisan, J.~J. Chung, R.~Siegwart, and J.~Alonso-Mora,
  ``Multi-agent path integral control for interaction-aware motion planning in
  urban canals,'' in \emph{2023 IEEE International Conference on Robotics and
  Automation (ICRA)}, 2023, pp. 1379--1385.

\bibitem{9802523}
A.~Romero, S.~Sun, P.~Foehn, and D.~Scaramuzza, ``Model predictive contouring
  control for time-optimal quadrotor flight,'' \emph{IEEE Transactions on
  Robotics}, vol.~38, no.~6, pp. 3340--3356, 2022.

\bibitem{8678822}
M.~Muehlebach and R.~D’Andrea, ``A method for reducing the complexity of
  model predictive control in robotics applications,'' \emph{IEEE Robotics and
  Automation Letters}, vol.~4, no.~3, pp. 2516--2523, 2019.

\bibitem{10777539}
J.~Ubbink, R.~Viljoen, E.~Aertbeliën, W.~Decré, and J.~De~Schutter, ``From
  instantaneous to predictive control: A more intuitive and tunable mpc
  formulation for robot manipulators,'' \emph{IEEE Robotics and Automation
  Letters}, vol.~10, no.~1, pp. 748--755, 2025.

\bibitem{GONZALEZGARCIA2022oa}
A.~Gonzalez-Garcia, I.~Collado-Gonzalez, R.~Cuan-Urquizo, C.~Sotelo, D.~Sotelo,
  and H.~Castañeda, ``Path-following and lidar-based obstacle avoidance via
  nmpc for an autonomous surface vehicle,'' \emph{Ocean Engineering}, vol. 266,
  p. 112900, 2022.

\bibitem{Aaron2014}
A.~D. Ames, J.~W. Grizzle, and P.~Tabuada, ``Control barrier function based
  quadratic programs with application to adaptive cruise control,'' in
  \emph{Proc. of 53rd IEEE Conference on Decision and Control}, 2014, pp.
  6271--6278.

\bibitem{xiao2021high}
W.~{Xiao} and C.~{Belta}, ``High-order control barrier functions,'' \emph{IEEE
  Transactions on Automatic Control}, vol.~67, no.~7, pp. 3655--3662, 2021.

\bibitem{Darweesh_2017jrm}
H.~Darweesh, E.~Takeuchi, K.~Takeda, Y.~Ninomiya, A.~Sujiwo, L.~Y. Morales,
  N.~Akai, T.~Tomizawa, and S.~Kato, ``Open source integrated planner for
  autonomous navigation in highly dynamic environments,'' \emph{Journal of
  Robotics and Mechatronics}, vol.~29, no.~4, pp. 668--684, 2017.

\bibitem{Huang2019}
Y.~Huang, L.~Chen, and P.~Gelder, ``Generalized velocity obstacle algorithm for
  preventing ship collisions at sea,'' \emph{Ocean Engineering}, vol. 173, pp.
  142--156, 2019.

\bibitem{Shi2019}
B.~Shi, Y.~Su, C.~Wang, L.~Wan, and Y.~Luo, ``Study on intelligent collision
  avoidance and recovery path planning system for the waterjet-propelled
  unmanned surface vehicle,'' \emph{Ocean Engineering}, vol. 182, pp. 489--498,
  2019.

\bibitem{Wiig2020}
M.~S. Wiig, K.~Y. Pettersen, and T.~R. Krogstad, ``Collision avoidance for
  underactuated marine vehicles using the constant avoidance angle algorithm,''
  \emph{IEEE Transactions on Control Systems Technology}, vol.~28, no.~3, pp.
  951--966, 2020.

\bibitem{DeVries2022RegulationsCanals}
J.~De~Vries, E.~Trevisan, J.~Van Der~Toorn, T.~Das, B.~Brito, and
  J.~Alonso-Mora, ``{Regulations Aware Motion Planning for Autonomous Surface
  Vessels in Urban Canals},'' in \emph{Proceedings - IEEE International
  Conference on Robotics and Automation}.\hskip 1em plus 0.5em minus
  0.4em\relax Institute of Electrical and Electronics Engineers Inc., 2022, pp.
  3291--3297.

\bibitem{Thirugnanam2022}
A.~Thirugnanam, J.~Zeng, and K.~Sreenath, ``Safety-critical control and
  planning for obstacle avoidance between polytopes with control barrier
  functions,'' in \emph{2022 International Conference on Robotics and
  Automation (ICRA)}, 2022, pp. 286--292.

\bibitem{Wei2024}
W.~Wang, W.~Xiao, A.~Gonzalez-Garcia, J.~Swevers, C.~Ratti, and D.~Rus,
  ``Robust model predictive control with control barrier functions for
  autonomous surface vessels,'' in \emph{2024 IEEE International Conference on
  Robotics and Automation (ICRA)}, 2024, pp. 6089--6095.

\bibitem{WeiICRA2018}
W.~Wang, L.~Mateos, S.~Park, P.~Leoni, B.~Gheneti, F.~Duarte, C.~Ratti, and
  D.~Rus, ``Design, modeling, and nonlinear model predictive tracking control
  of a novel autonomous surface vehicle,'' in \emph{Proc. 2018 IEEE Int. Conf.
  Robot. Autom}, 2018, pp. 6189--6196.

\bibitem{BOS20234877}
M.~Bos, B.~Vandewal, W.~Decré, and J.~Swevers, ``Mpc-based motion planning for
  autonomous truck-trailer maneuvering,'' \emph{IFAC-PapersOnLine}, vol.~56,
  no.~2, pp. 4877--4882, 2023, 22nd IFAC World Congress.

\bibitem{gillis2020effortless}
J.~Gillis, B.~Vandewal, G.~Pipeleers, and J.~Swevers, ``Effortless modeling of
  optimal control problems with rockit,'' in \emph{39th Benelux Meeting on
  Systems and Control}, vol. 138.\hskip 1em plus 0.5em minus 0.4em\relax
  Elspeet, The Netherlands, 2020.

\bibitem{vanroye2023fatrop}
L.~Vanroye, A.~Sathya, J.~De~Schutter, and W.~Decr{\'e}, ``Fatrop: A fast
  constrained optimal control problem solver for robot trajectory optimization
  and control,'' in \emph{2023 IEEE/RSJ International Conference on Intelligent
  Robots and Systems (IROS)}.\hskip 1em plus 0.5em minus 0.4em\relax IEEE,
  2023, pp. 10\,036--10\,043.

\bibitem{ZHANG2024229}
Y.-Y. Zhang, J.~Billet, and P.~Slaets, ``Experimental identification of
  decoupled ship dynamic models for an autonomous catamaran urban cargo
  vessel,'' \emph{IFAC-PapersOnLine}, vol.~58, no.~20, pp. 229--234, 2024.

\bibitem{Ferreau2014}
H.~Ferreau, C.~Kirches, A.~Potschka, H.~Bock, and M.~Diehl, ``{qpOASES}: A
  parametric active-set algorithm for quadratic programming,''
  \emph{Mathematical Programming Computation}, vol.~6, no.~4, pp. 327--363,
  2014.

\bibitem{TixiaoIROS2020a}
T.~{Shan}, B.~{Englot}, D.~{Meyers}, W.~{Wang}, C.~{Ratti}, and D.~{Rus},
  ``{LIO-SAM}: Tightly-coupled lidar inertial odometry via smoothing and
  mapping,'' in \emph{2020 IEEE/RSJ International Conference on Intelligent
  Robots and Systems (IROS)}, 2020, pp. 5135--5142.

\bibitem{opencv_library}
G.~Bradski, ``{The OpenCV Library},'' \emph{Dr. Dobb's Journal of Software
  Tools}, 2000.

\end{thebibliography}
\end{document}